\documentclass[10pt,twocolumn,letterpaper]{article}

\usepackage[pagenumbers]{cvpr}      
\usepackage[super]{nth}
\usepackage[dvipsnames]{xcolor}

\usepackage{graphicx}
\usepackage{amsmath}
\usepackage{amssymb}
\usepackage{booktabs}
\usepackage[utf8]{inputenc} 
\usepackage[T1]{fontenc}    
\usepackage{url}            
\usepackage{booktabs}       
\usepackage{amsfonts}       
\usepackage{nicefrac}       
\usepackage{microtype}      
\usepackage{xcolor}         
\usepackage{graphicx}
\usepackage{amsmath}
\usepackage{amssymb}
\usepackage{makecell}
\usepackage{engord}
\usepackage[normalem]{ulem}
\usepackage{adjustbox}
\usepackage{enumitem}

\usepackage{ifthen}

\definecolor{FanColor}{rgb}{0.8,0,0.8}
\newcommand{\fan}[1]{{\color{FanColor}[Fan: #1]}}

\definecolor{XingjiaColor}{rgb}{0.0,0.1,0.9}
\newcommand{\xingjia}[1]{{\color{XingjiaColor}[Xingjia: #1]}}

\newcommand{\warning}[1]{{\it\color{red} #1}}
\newcommand{\toremove}[1]{{\it\color{red} (To remove) #1}}
\newcommand{\note}[1]{{\it\color{blue} #1}}
\newcommand{\nothing}[1]{}

{
\renewcommand{\fan}[1]{}
\renewcommand{\xingjia}[1]{}

\renewcommand{\warning}[1]{}
\renewcommand{\toremove}[1]{}
\renewcommand{\note}[1]{}
\renewcommand{\nothing}[1]{}
}{}

\definecolor{cvprblue}{rgb}{0.21,0.49,0.74}
\usepackage[pagebackref,breaklinks,colorlinks,citecolor=cvprblue]{hyperref}
\usepackage[capitalize]{cleveref}
\def\paperID{3555} 
\def\confName{CVPR}
\def\confYear{2024}

\title{$\mathcal{Z}^*$: \uline{Z}ero-shot \uline{S}tyle \uline{T}ransfer via \uline{A}ttention \uline{R}earrangement}
\author{Yingying Deng$^1$$^*$,$\qquad$ Xiangyu He$^1$\thanks{These authors contributed equally.},$\qquad$ Fan Tang$^2$, $\qquad$Weiming Dong$^1$\\
Institute of Automation, Chinese Academy of Science$^1$\\
Institute of Computing Technology, Chinese Academy of Science$^2$\\
{\tt\small \{dengyingying2017, weiming.dong\}@ia.ac.cn, xiangyu.he@nlpr.ia.ac.cn, tangfan@ict.ac.cn}
}

\begin{document}

\maketitle

\begin{figure*}
    \captionsetup{type=figure}
    \includegraphics[width= \linewidth]{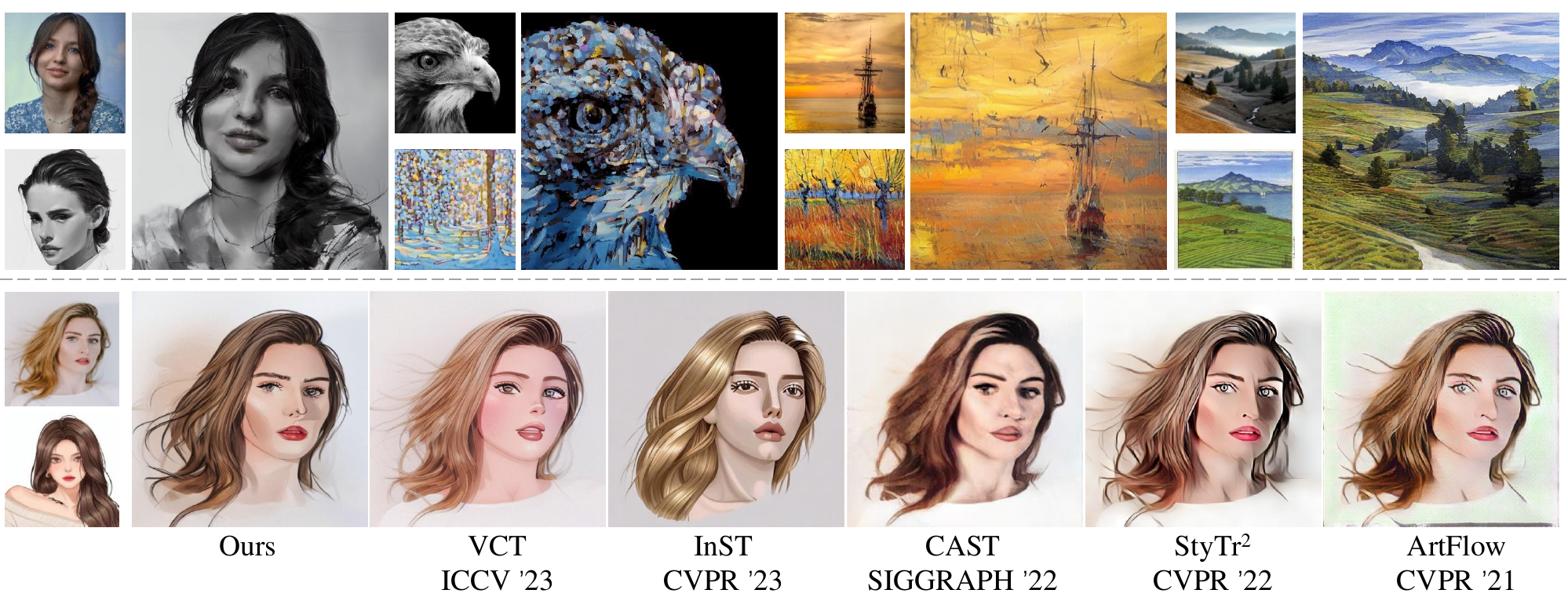}
    \caption{Image style transfer results by the proposed $\mathcal{Z}^*$.
\textbf{Top}: The stylized results by style/content references of different types. Our method can well balance the contents and styles in the results.
\textbf{Bottom}: Comparisons with state-of-the-art methods, including diffusion-based models (VCT~\cite{cheng:2023:ict} and InST~\cite{zhang:2023:inversion}), transformer-based model (StyTr$^2$~\cite{Deng:2022:CVPR}), flow-based model (ArtFlow~\cite{an:2021:artflow}), and CNN-based model (CAST~\cite{zhang:2022:domain}). Our method excels in generating stylized images with vivid style patterns and accurate content details.}
    \label{fig:position}
    \label{fig:teaser}
    \end{figure*}%

\begin{abstract}
\vspace{-14pt}

Despite the remarkable progress in image style transfer, formulating style in the context of art is inherently subjective and challenging. 
In contrast to existing learning/tuning methods, this study shows that vanilla diffusion models can directly extract style information and seamlessly integrate the generative prior into the content image without retraining.
Specifically, we adopt dual denoising paths to represent content/style references in latent space and then guide the content image denoising process with style latent codes.
We further reveal that the cross-attention mechanism in latent diffusion models tends to blend the content and style images, resulting in stylized outputs that deviate from the original content image.
To overcome this limitation, we introduce a cross-attention rearrangement strategy.
Through theoretical analysis and experiments, we demonstrate the effectiveness and superiority of the diffusion-based \uline{z}ero-shot \uline{s}tyle \uline{t}ransfer via \uline{a}ttention \uline{r}earrangement, $\mathcal{Z}$-STAR.
\end{abstract}

\section{Introduction}
\label{sec:Introduction}
The task of image style transfer has received significant attention in the research community, with numerous machine learning techniques utilized, such as convolutional neural networks (CNN)~\cite{gatys:2016:image, Huang:2017:Arbitrary, johnson:2016:perceptual, sheng:2018:avatar, park:2019:arbitrary, deng:2020:arbitrary}, flow-based networks~\cite{an:2021:artflow}, visual transformers (ViT)~\cite{wu:2021:styleformer,Deng:2022:CVPR}, and diffusion models~\cite{zhang:2023:inversion,cheng:2023:ict}.
After completing the training process, the output stylized image is generated based on a content and style image, as shown in Figure~\ref{fig:position}.
The generated image retains the content layout from the input content while adopting a similar style to that of the input style.
In essence, the networks are trained by ensuring that the generated image and the content/style image exhibit content/style similarity.
%

While certain methods, such as \cite{Deng:2022:CVPR,an:2021:artflow,Huang:2017:Arbitrary}, employ the Gram matrix~\cite{gatys:2016:image} to measure global style similarity and achieve promising style transfer outcomes, the second-order statistics contained in the Gram matrix are limited in their ability to capture intricate style patterns and fail to transfer corresponding local features from the content image to the style image (e.g., hair and eyes in StyTr$^2$~\cite{Deng:2022:CVPR}, ArtFlow~\cite{an:2021:artflow}, AdaIN~\cite{Huang:2017:Arbitrary} in Figure~\ref{fig:position}). 
To address this issue, CAST~\cite{zhang:2022:domain} proposes a contrastive loss that leverages the relationships between positive and negative examples to encourage the result to conform to the distribution of styles. 
However, CAST also faces the obstacle of generating stylized outcomes with vivid, fine-grained style details. 
In reality, \textit{the contours and forms of a painting should be subject to the adaptable preferences of the artist's painting techniques, rather than being rigidly determined by the content and style images}. 
In light of this, we rethink the role of training in style transfer and reveal that the generative model used to describe the distribution of images has already learned the art of transfer. 

With the advent of diffusion models, text-controlled image editing and translation have gained unprecedented attention. 
When given an image as input, the diffusion model can generate an artistic image incorporating a style-related prompt.
However, the textual prompt is often too coarse to effectively express the desired style details.
While prior approaches such as InST~\cite{zhang:2023:inversion} and VCT~\cite{cheng:2023:ict} endeavor to employ an image-controlled diffusion model for image style transfer and translation, they necessitate training a style embedding for each input style, leading to challenges in distilling precise style representations and resulting in deviations from the input style while failing to preserve content (see InST and VCT results in Figure~\ref{fig:position}).
In contrast to \cite{zhang:2023:inversion, cheng:2023:ict}, where control information is encoded as text embedding, we propose that the vanilla diffusion model is capable of extracting style information directly from the desired style image and fusing it into the content image \textit{without requiring re-training or tuning}.

In this paper, we leverage the prior knowledge from latent diffusion~\cite{rombach:2022:sd} and propose a \uline{z}ero-shot (\textit{i.e.,} training-free) \uline{s}tyle \uline{t}ransfer method via \uline{a}ttention \uline{r}earrangement, namely $\mathcal{Z}$-\textit{STAR} ($\mathcal{Z}^*$), to addresses the issues above. 
To obtain generative image priors, we employ dual diffusing paths to invert the style and content images. 
The features obtained from diffusion models naturally represent content and style information and could be fused by the attention mechanism.
However, without a training process, it is challenging to strike a balance between content and image influence.
That is to say, the naive cross-attention operations are not optimal for directly integrating content and style latent status in the denoising process.
Content structure may be compromised due to inaccurate cross-attention values (see Sec.~\ref{sec:method.attre} for more details).
Therefore, we propose a multi-cross attention rearranged strategy that manipulates content and style information from images and seamlessly fuses them in the diffusion latent space.
By leveraging a tailored attention mechanism, the diffusion model can naturally address the constraints of content and style without necessitating additional supervision.
Experimental results demonstrate that our method generates satisfactory results with well-preserved content and vivid styles adapted to content structures.
In summary, our main contributions are as follows:
\begin{itemize}

\item  A zero-shot image style transfer method leveraging the generative prior knowledge to conduct image stylization without retraining/tuning.
\item  A rearranged attention mechanism to disentangle and fuse content/style information in the diffusion latent space.
\item  Various experiments demonstrate that our method can generate outstanding style transfer results, naturally fusing and balancing content and style from two input images.
\end{itemize}

\section{Related Work}
\label{sec:related_work}

\paragraph{Image style transfer.}
Since Gatys~\etal~\cite{gatys:2016:image}  discovered that perceptual features can effectively represent content and Gram matrices may express style in CNNs, various frameworks for style transfer trained by content and style loss have been proposed.
CNN-based methods~\cite{li:2016:precomputed,Huang:2017:Arbitrary,deng:2020:arbitrary,deng:2021:arbitrary,sheng:2018:avatar,2019:An:Ultrafast,risser:2017:stable,2019:Lu:A,2020:Wang:Collaborative} achieved success by exploring the fusion of content and style representation.
Some works~\cite{Deng:2022:CVPR, wu:2021:styleformer,wei:2022:comparative,Tang_2023_CVPR,wang2022fine,zhang2022s2wat} utilize the long-range feature represention ability of Transformer~\cite{xu:2022:transformers} and enhance stylization effects.
However, the Gram matrix measures second-order statistics of the entire image, which may not be sufficient for style representation.
Recent works~\cite{zhang:2022:domain,chen2021artistic,yang2023zero} use contrastive loss to replace style loss based on the Gram matrix, which is effective in processing fine-detail style patterns.
Despite the continuous progress of existing methods, precise style representation remains challenging, and inaccurate style expression may lead to unsatisfactory stylized results.
In light of this challenge, we aim to develop a zero-shot style transfer method that does not rely on explicit style constraints.

\paragraph{Diffusion for image generation.}

Diffusion models have demonstrated impressive results in text-to-image generation~\cite{rombach:2022:sd,Ramesh:dalle2:2012,Saharia:imagen:2022,nichol2021glide} and image editing~\cite{Bahjat:imagic:2022,Hertz:p2p:2023,Mokady:nulltext:2022, tecent:masactrl:2023,Tumanyan:plugplay:2022,zhang2023sine,couairon2022diffedit,wu2023latent,brooks2023instructpix2pix,avrahami2022blended,pan2023effective}. 
However, certain methods like Imagic~\cite{Bahjat:imagic:2022} require fine-tuning the entire diffusion model for each instruction, which can be time-consuming and memory-intensive.
To address this, Prompt-to-prompt~\cite{Hertz:p2p:2023} introduces cross-attention maps during the diffusion process by replacing or reweighting the attention map between text prompts and edited images. 
Additionally, NTI~\cite{Mokady:nulltext:2022} proposes null-text optimization based on Prompt-to-prompt to enable real image editing.
In order to reduce reliance on text prompts, StyleDiffuison~\cite{li2023stylediffusion} incorporates a mapping network to invert the input image to a context embedding, which is then utilized as a key in the cross-attention layers. However, manipulating the cross-attention solely between text and image can be challenging for achieving precise control.
To address this, Plug-and-Play~\cite{Tumanyan:plugplay:2022} and MasaCtrl~\cite{tecent:masactrl:2023}  focus on spatial features using self-attention in the U-Net of the latent diffusion model.
While these methods can accomplish text-guided style transfer by inputting a text prompt like ``a pencil drawing'', simple words may not be sufficient to describe fine-detail style patterns.
To address this limitation, InST~\cite{zhang:2023:inversion} and VCT~\cite{cheng:2023:ict} employ an inversion-based image style transfer/translation scheme that can train a style image into a style embedding to guide the generated results.
In this paper, we demonstrate that style images alone (\textit{i.e.}, without pseudo-text guidance) are adequate for latent diffusion models to achieve image-guided style transfer, without requiring additional training.

\section{Preliminary}
\label{sec:preliminary}
\noindent \textbf{Attention Mechanism}~\cite{BahdanauCB14} was introduced as a powerful tool in neural network architectures for aggregating information and later adopted by Vaswani et al. \cite{VaswaniSPUJGKP17} as a fundamental building block for machine translation:
\begin{equation}
    \text{Attention}(Q,K,V)=\text{Softmax}(\frac{QK^T}{\sqrt{d}})V.
\end{equation}
Attention-based vision transformers \cite{dosovitskiy2021vit, liu:2021:SwinTransformer}have demonstrated remarkable empirical results on mainstream benchmarks, solidifying attention mechanisms as a crucial component in modern deep neural networks.
Moreover, by incorporating additional information, along with the utilization of \textit{Key} and \textit{Value} vectors, cross-attention has proved effective in latent diffusion models for applying conditions to the denoising process.

\noindent \textbf{Diffusion Model}, as described in the literature, belongs to a class of generative models that employ Gaussian noise to generate desired data samples. 
This is accomplished through an iterative process of noise removal, where a forward process is defined to add noise to an initial data sample $x_0$, resulting in a noisy sample $x_t$ at time-step $t$, according to a predetermined noise-adding schedule $\alpha_t$:
\begin{equation}
    x_t=\sqrt{\alpha_t}\cdot x_0+\sqrt{1-\alpha_t}\cdot z,\text{ } z\sim\mathcal{N}(0,\mathbf{I}).
\label{forward_process}
\end{equation}
Additionally, a corresponding reverse process is also defined:
\begin{equation}
    x_{t-1} = \frac{1}{\sqrt{\alpha_t}} \Big( \mathbf{x}_t - \frac{1 - \alpha_t}{\sqrt{1 - \bar{\alpha}_t}} \boldsymbol{\epsilon}_\theta(\mathbf{x}_t, t) \Big) + \sigma_t z.
\end{equation}
The backward process aims to gradually denoise $x_T\sim \mathcal{N}(0,\mathbf{I})$, where a cleaner image $x_{t-1}$ is obtained at each step. This is accomplished by a neural network $\boldsymbol{\epsilon}_\theta(x_t, t)$ that predicts the added noise $z$. 

Utilizing a U-Net integrated with an attention mechanism as $\boldsymbol{\epsilon}_\theta(x_t, t)$ is a common approach.
This configuration allows for self-attention to capture long-range interactions among image features, while cross-attention receives a guiding signal from the given text prompt.
The attention mechanisms are formulated as: 
\begin{equation}
    f_t^l=\text{Attention}(Q_t^l,K_t^l,V_t^l).
\label{attention_form}
\end{equation}
Even as $K$ey and $V$alue at $l$-th layer may vary from image spatial or text features, they still adhere to the standard format.
\section{Method}
\label{sec:Method}
\begin{figure*}
\setlength{\abovecaptionskip}{2mm}
\centering
\includegraphics[width=0.98\linewidth]{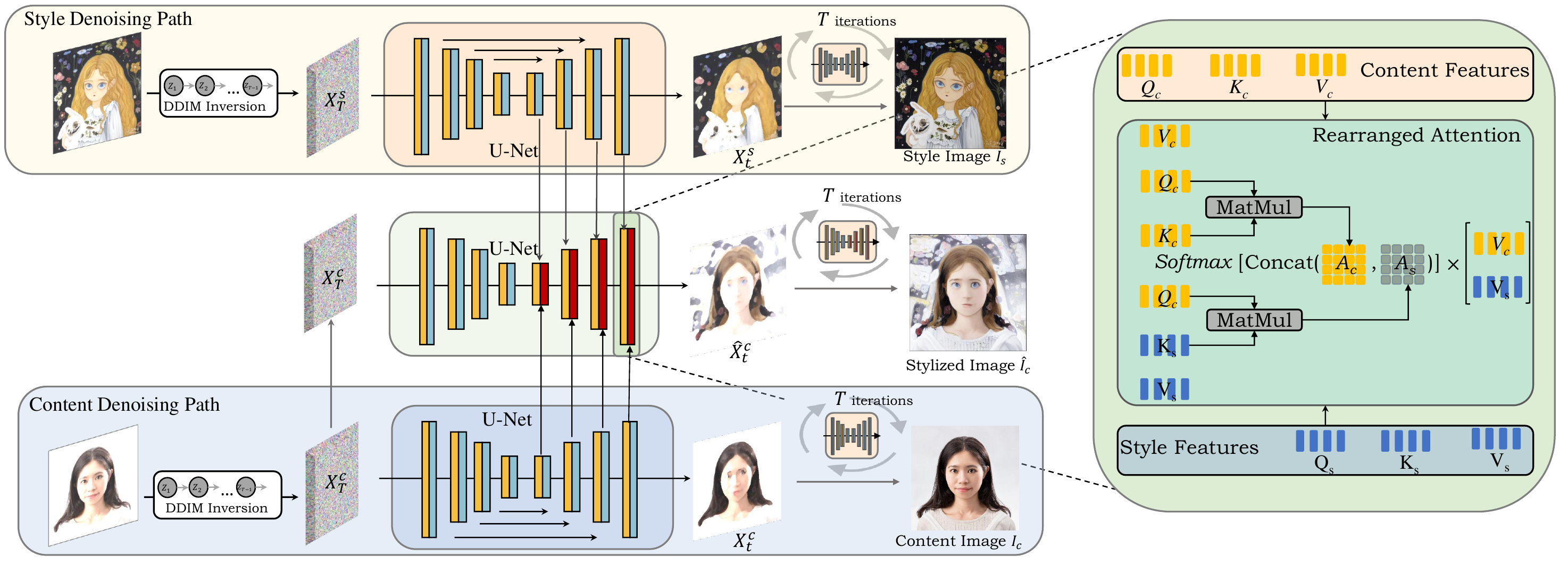}
\caption{Overall pipeline of our style transfer framework. The stylization process operates in the latent space. We perform DDIM inversion separately for the content and style images. During the denoising process, our Cross-attention Rearrangement is employed to integrate style patterns into the content structure. By iteratively performing 50 denoising steps, we are able to achieve the final stylized output. 
}
\label{fig:network}
\vspace{-10pt}
\end{figure*}

Our research is based on the observation that the attention module within the stable diffusion can effectively align features $K$ and $V$ with the query $Q$. 
Previous studies, such as \cite{tecent:masactrl:2023, Tumanyan:plugplay:2022}, have leveraged the self-attention layer to extract information from key and value features, which represent spatial attributes in the DDIM inversion process, for image editing applications.
However, in style transfer tasks, the simultaneous preservation of both style and content is crucial.
Consequently, we need to address two important questions to practically apply this observation:
\begin{itemize}
\item 
How can appropriate style features $K$ and $V$ be obtained for the stable diffusion model without necessitating re-training? It is crucial to emphasize that our objective is to directly extract style information from images rather than depending on a proxy prompt like text embedding, which may lack the necessary level of detail.
\item Mere utilization of style attention~\cite{park2019arbitrary} results in poor content preservation under diffusion models without re-training. We aim to recreate the content image in a way that resembles the style image in terms of its contours, forms, and overall visual appearance.
\end{itemize}
To address these challenges, our approach adopts a two-fold strategy. Firstly, dual-path networks are designed to generate suitable $K$ and $V$ features during the reverse process. Secondly, we incorporate attention rearrangement techniques to better align content features with style features.

\subsection{Dual-path Networks}

In our proposed approach, we address a key limitation in the conventional stable diffusion model, where the text embedding remains unchanged throughout the reverse process, from timestamp $t\in[0,T]$, despite the desirable need for the style feature to adapt to the denoising stylized image. This adaptation is crucial as the initial denoising process involves the reconstruction of the image's shape and color, followed by the refinement of details such as contours and brushstrokes towards the end.

To address this limitation, we introduce a novel dual-path scheme that simultaneously generates the denoised style image and the stylized content image at the same timestamp $T$. This is achieved through the following equations:
\begin{equation}
  I_s=\mathcal{G}_\theta(\epsilon_{I_s}, \{f_s\}, T),\quad I_c=\mathcal{G}_\theta(\epsilon_{I_c}, \{f_c\}, T).
\end{equation}
This ensures that the features in both networks are naturally aligned in the time dimension. Specifically, given a content image $I_c$ and a style image $I_s$, our objective is to obtain a stylized result $\hat{I_c}$ that retains the content of $I_c$ while incorporating stylistic patterns from $I_s$. This is achieved through the following equation:
\begin{equation}
  \hat{I_c}=\mathcal{G}_\theta(\epsilon_{I_c}, \{f_s, f_c\}, T),
  \label{network_arch}
\end{equation}
where $\mathcal{G}_\theta(\cdot,\cdot,T)$ represents the application of denoising for $T$ steps in the diffusion model, using fixed pre-trained weights $\theta$. The term $\epsilon_{I_*}$ denotes the noisy $x_T$ generated in the forward process by progressively adding Gaussian noise to $I_c$ or $I_s$, as described in Eq.~(\ref{forward_process}). The notation $\{f_s, f_c\}$ refers to the spatial U-Net features in the diffusion models from the style and content images, respectively, which are utilized in cross-attention.

As shown in Figure~\ref{fig:network}, we utilize ddim inversion to invert the style image and content image, obtaining the diffusion trajectories ${x}_{[0:T]}^c$ and ${x}_{[0:T]}^s$. Subsequently, we introduce a novel cross attention arrangement to disentangle and fuse content and style information (i.e., $f_c$ and $f_s$, denoted as $Q$uery, $K$ey and $V$alue), within the diffusion latent space using U-Net at each timestamp $t$.
Through $T$ denoising steps, we convert the stylized latent features $\hat{f}_{c}$, generated by the rearranged attention, into the style transfer result $\hat{I}_{c}$.

\subsection{Attention Rearrangement}
\label{sec:method.attre}
\begin{figure}
\setlength{\abovecaptionskip}{2mm}
\centering
\includegraphics[width= \linewidth]{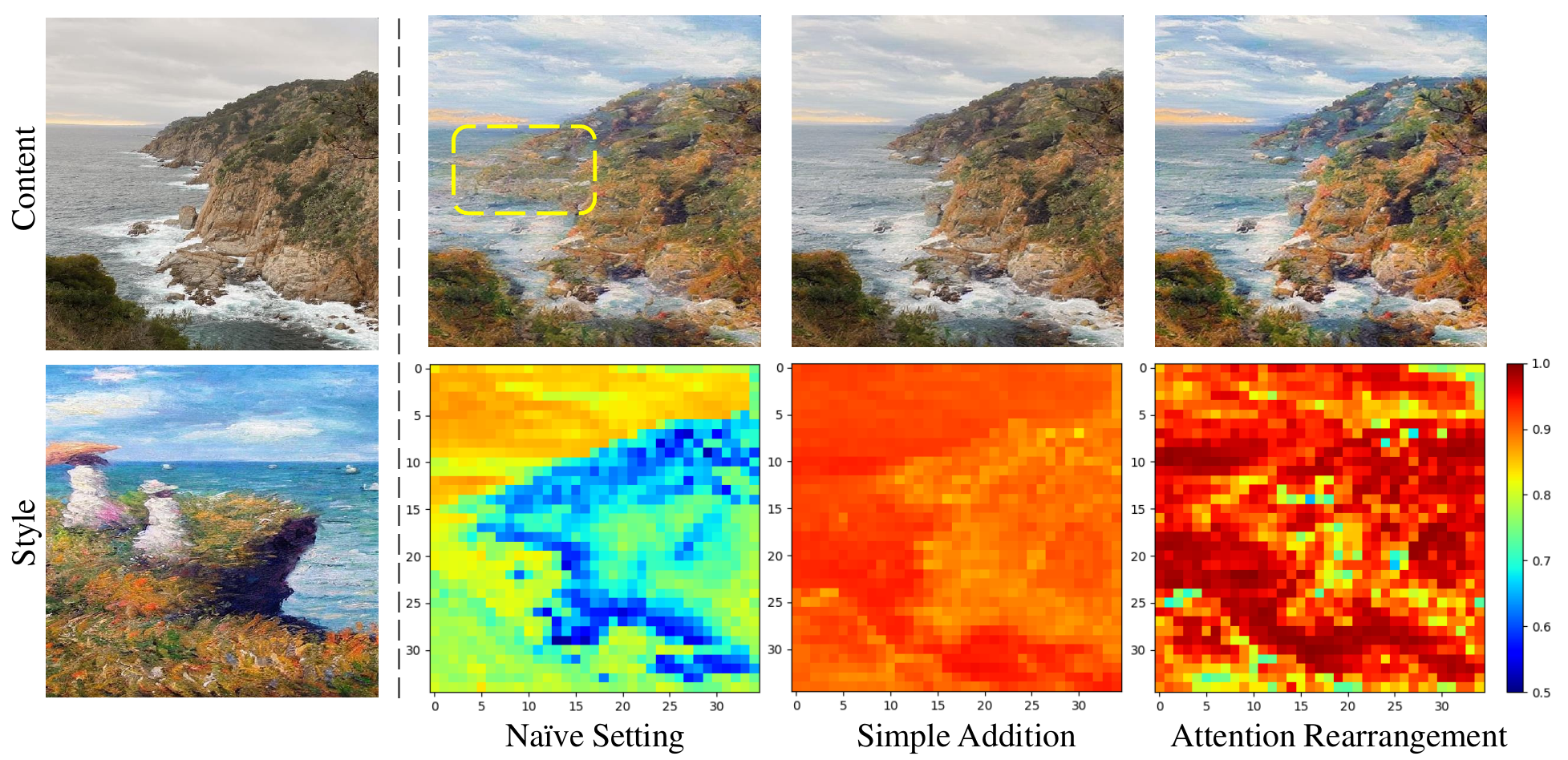}
\caption{The outcome of the ``Naive Setting'' exhibits a bias towards emphasizing the style patterns while neglecting the original content structures. The distorted area corresponds to the low similarity portion between self-attention and cross-attention results, suggesting that the $V_s$ fails to adequately reconstruct the target region, resulting in a loss of content. ``Simple Addition'' preserves an excessive number of content features, whereas the re-arranged attention achieves a more favorable trade-off.}
\label{fig:similarity}
\vspace{-10pt}
\end{figure}

As demonstrated in Eq.~(\ref{network_arch}), our attention mechanism incorporates two types of attention calculation between $f_s$ and $f_c$: style-cross attention for merging content and style features, and content self-attention for preserving structure. Since we use the standard self-attention, we mainly discuss the proposed style-cross attention in this section.
\paragraph{Naive Setting}
It is intuitive to represent content information, such as image structure, using the $Q$uery, and represent style information, such as color, texture, and object shape, using the $K$ey and $V$alue features. The style-cross attention then uses the content features to query the information from style images that best suit the input patch. Formally, the inputs of style-cross attention are features from the content latent space $c$ and style latent space $s$ where 
\begin{equation}
    \hat{f}_c=\text{Attn}(Q_c,K_s,V_s)=\text{Softmax}(\frac{Q_cK_s^T}{\sqrt{d}})V_s.
\label{eq:naive_setting}
\end{equation}
Despite the simplicity, we observe that the naive fusion setting in Eq.~(\ref{eq:naive_setting}) tends to prioritize the style patterns at the expense of the original content structures. Figure \ref{fig:similarity} displays the heatmap representing the cosine similarity between the results of cross-attention $\text{Atten}(Q_c,K_s,V_s)$ and self-attention $\text{Atten}(Q_c,K_c,V_c)$. It is observed that regions with low similarity scores correspond to pixels that have experienced a loss of content information.

\begin{figure}[t]
\begin{minipage}[t]{.15\textwidth}
  \centering
  \includegraphics[width=1.1\linewidth]{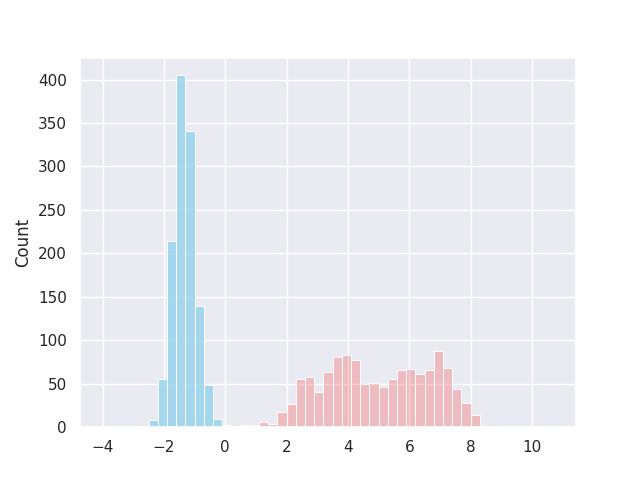}
  \caption*{(a) $\vec{q_c}K_s$}
\end{minipage}
\begin{minipage}[t]{.15\textwidth}
  \centering
  \includegraphics[width=1.1\linewidth]{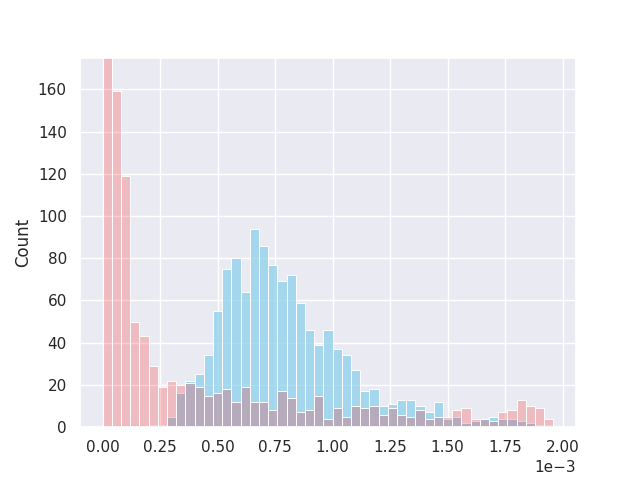}
  \caption*{(b) Softmax$(\vec{q_c}K_s)$}
\end{minipage}
\begin{minipage}[t]{.15\textwidth}
  \centering
  \includegraphics[width=1.1\linewidth]{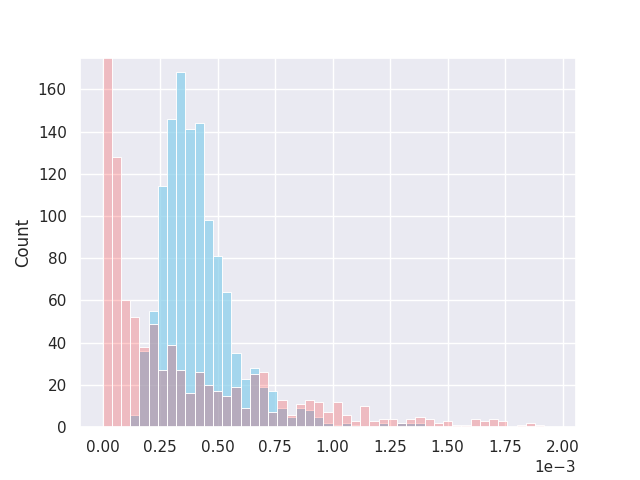}
  \caption*{(c) Ours}
\end{minipage}
\caption{Visualization of the distribution of $\vec{q_c}K_s$ for two content feature points (represented by blue and red bars) before and after the Softmax in (a) and (b). In (c), we display the distribution of $\vec{q_c}K_s$ normalized by Eq.(\ref{ours_form}). It is observed that Softmax operation tends to overly amplify smaller values of $\vec{q_c}K_s$ (e.g., blue bars being shifted further to the right compared to the red bars in (b), even though the original $\vec{q_c}K_s$ values are mostly less than 0). In contrast, the negatively correlated values are limited to small magnitudes before and after normalization in (c).}
\vspace{-3mm}
\label{fig:histogram}
\end{figure}

\paragraph{Simple Addition}
To tackle the aforementioned issue, we propose a simple solution by enhancing the content information in $\hat{f}_c$ through the reintroduction of content self-attention. The equation is formulated as follows:
\begin{equation}
   \hat{f}_c=\lambda\cdot\text{Attn}(Q_c,K_s,V_s)+(1-\lambda)\cdot\text{Attn}(Q_c,K_c,V_c),
   \label{cross_attention_naive_add}
\end{equation}
where $\lambda\in[0,1]$.
However, we found the selection of $\lambda$ can be delicate. For instance, certain content pixels exhibit a weak correlation with style, as indicated by small values of $\vec{q}_cK_s^T$ (represented by blue bars in Figure~\ref{fig:histogram}(a), where $\vec{q}_cK_s^T$ is less than 0). It is desirable to assign smaller attention weights to these pixels in order to minimize their negative impact. Conversely, in other scenarios where $\vec{q}_cK_s^T$ yields meaningful values (represented by red bars in Figure~\ref{fig:histogram}(a)), we aim to assign larger attention weights to these pixels. Unfortunately, due to the inherent nature of the Softmax function, which disregards the absolute magnitudes and solely amplifies the differences between $\vec{q}_cK_s^T$ values, we observe counter-intuitive results as depicted in Figure~\ref{fig:histogram}(b), where smaller $\vec{q}_cK_s^T$ values result in larger attention weights after Softmax normalization. In such cases, we need to introduce an additional variable, denoted as $\lambda$, to compensate for this deficiency. However, it is worth noting that a pre-defined $\lambda$ value cannot cater to the requirements of every content/style image pair.

\paragraph{Cross-attention Rearrangement}
Though a handcraft $\lambda$ outside of $\text{Atten}(\cdot,\cdot,\cdot)$ can not adapt to input images, we find that it can be achieved with an $\lambda$ inside $\text{Softmax}(\cdot)$. By expressing Eq.~(\ref{cross_attention_naive_add}) in matrix form through an equivalent reformulation, we obtain:
\begin{align}
   \hat{f}_c&=\left[
   \begin{matrix}
       \lambda\cdot\sigma(\frac{Q_cK_s^T}{\sqrt{d}}), & (1-\lambda)\cdot\sigma(\frac{Q_cK_c^T}{\sqrt{d}})
   \end{matrix}\right] *\left[
   \begin{matrix}
       V_s\\
       V_c
       \end{matrix}\right]\label{former_atten_form}\\
   &=A*V'^T.
\end{align}
Here, $\sigma(\cdot)$ represents the Softmax function, and each row in $A\in\mathbb{R}^{N\times 2N}$, denoted as $\vec{a}\in\mathbb{R}^{2N}$, is normalized, \text{i.e.,} $\vec{a}\cdot \vec{\mathbf{1}}^T=1$. This normalization inspires us to re-construct the matrix $A$ in the form of applying Softmax to rows, i.e.,
\begin{align}
    A'&=\sigma(\left[
   \begin{matrix}
       \lambda\cdot\frac{Q_cK_s^T}{\sqrt{d}}, & \frac{Q_cK_c^T}{\sqrt{d}}
   \end{matrix}\right])\label{ours_form}\\
\hat{f}_c'&=A'*V'^T=\sigma(\left[
   \begin{matrix}
       \lambda\cdot\frac{Q_cK_s^T}{\sqrt{d}}, & \frac{Q_cK_c^T}{\sqrt{d}},
   \end{matrix}\right]) * \left[
   \begin{matrix}
       V_s\\
       V_c
\end{matrix}\right].
\label{lambda}
\end{align}
In contrast to the previous attention formulation presented in Eq.~(\ref{former_atten_form}), the newly proposed rearranged attention matrix $A'\in\mathbb{R}^{N\times 2N}$ takes into account both the intra-content feature differences and the inter-content and style feature differences simultaneously during the application of the $\text{Softmax}(\cdot)$ function for output normalization. The rearranged attention matrix effectively enhances significant values of both $\vec{q}_cK_s^T$ and $\vec{q}_cK_s^T$ at each pixel, while automatically suppressing the small values of $\vec{q}_cK_s^T$ when the content pixel corresponding to $\vec{q}_c$ is irrelevant to all style pixels.

\paragraph{Superiority of Cross-attention Rearrangement}
The Cross-attention Rearrangement can be considered as a more versatile formulation. Its properties can be demonstrated as follows: (i) In cases where the correlation between the style and content images is weak, i.e., when each element $\vec{q}_c\vec{k_s}^T$ in $Q_cK_s^T$ approaches $-\infty$, the modified attention $\hat{f}_c'=A'*V'^T$ reduces to the standard self-attention of content images, denoted as $\text{Attention}(Q_c,K_c,V_c)$. (ii) When the correlation between the style and content images is strong, if the maximum value of $\vec{q}_c\vec{k_s}^T$ is approximately equal to the maximum value of $\vec{q}_c\vec{k_c}^T$, and the Softmax operation generates an approximate one-hot probability distribution, then $\hat{f}_c'=A'*V'^T$ is equivalent to Eq.~(\ref{cross_attention_naive_add}). (iii) Last but not least, Eq.~(\ref{cross_attention_naive_add}) can be rewritten using $A'$ as follows\footnote{Proof can be found in the Appendix's Section 3.}:
\begin{align}
    \hat{f}_c&=\frac{1}{2}\cdot\text{Attn}(Q_c,K_s,V_s)+\frac{1}{2}\cdot\text{Attn}(Q_c,K_c,V_c),\\
    &=\sigma(\left[
   \begin{matrix}
       \frac{Q_cK_s^T}{\sqrt{d}}+C, & \frac{Q_cK_c^T}{\sqrt{d}}
   \end{matrix}\right])*\left[
   \begin{matrix}
       V_s\\
       V_c
   \end{matrix}\right],
   \label{naive_add_new_form}
\end{align}
where
\begin{equation}
    C=\ln{\frac{\sum_j\exp{\big([Q_cK_c^T]_{\cdot,j}\big)}}{\sum_j\exp{\big([Q_cK_s^T]_{\cdot,j}\big)}}}.
\end{equation}
In Eq.~(\ref{naive_add_new_form}), the simple addition of self-attention to cross-attention introduces an additional term $C$. This variable serves to magnify all elements within $Q_cK_s^T$, including small values that represent weak correlations between style and content features, which are deemed inconsequential and should be disregarded. Consequently, the incorporation of $C$ may introduce an increased level of noise into the Softmax function, thereby resulting in sub-optimal outcomes.

\paragraph{Conditional Control}

\begin{figure}
\setlength{\abovecaptionskip}{2mm}
\centering
\includegraphics[width= \linewidth]{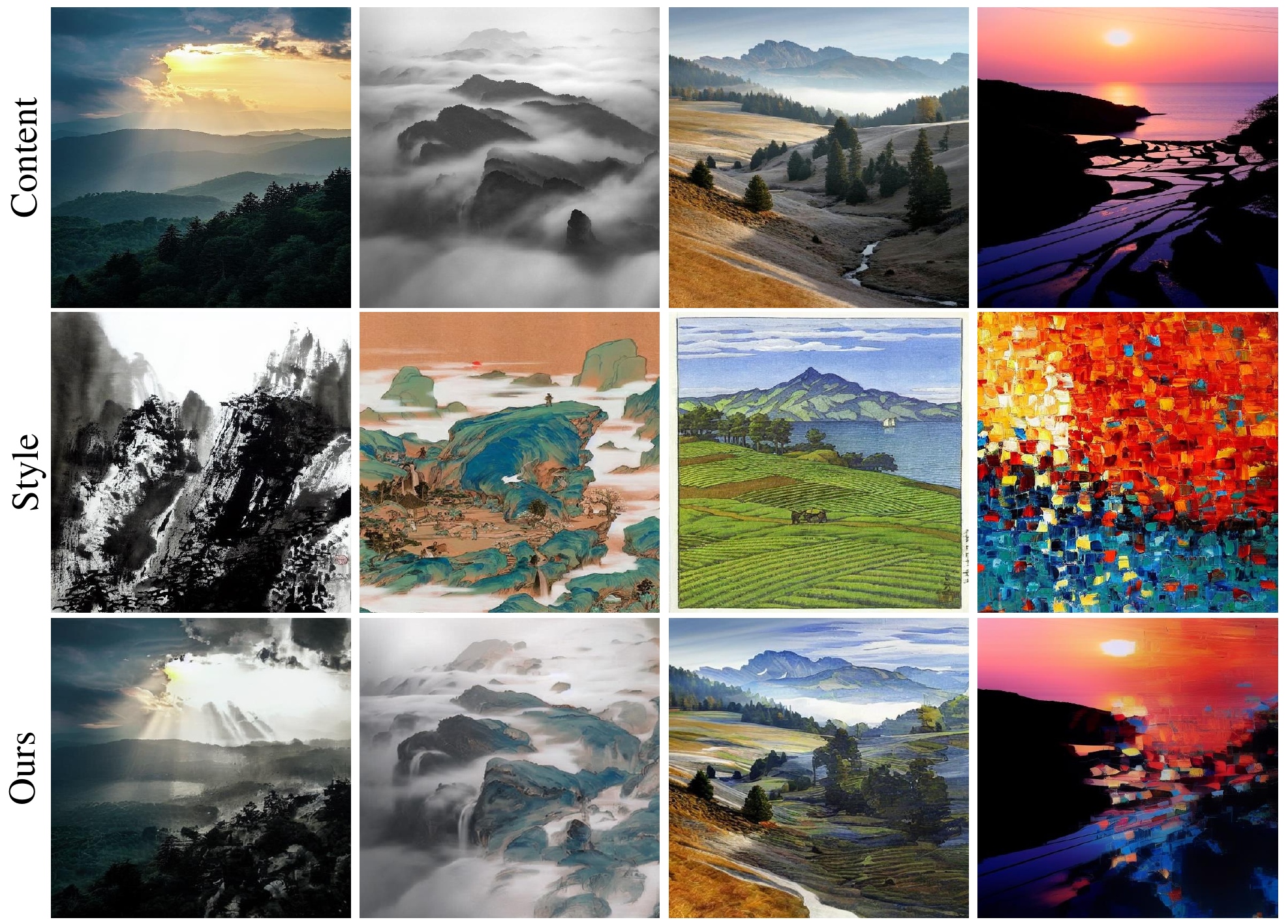}
\caption{Horizontal gradient effect of style created by our method. Image style gradually strengthens from left to right.}
\label{fig:style_grad}
\vspace{-10pt}
\end{figure}

The simplicity of Eq.~(\ref{ours_form}) allows for easy extension of the attention rearrangement technique to more complex downstream applications. To illustrate this, we introduce an additional mapping function $\phi(\cdot)$ on $\frac{Q_cK_c^T}{\sqrt{d}}$, which provides enhanced control over a specific region $\Omega$ for image style transfer. The modified equation is given by:
\begin{equation}
A'=\sigma(\left[\begin{matrix}
       \phi(\frac{Q_cK_s^T}{\sqrt{d}}), & \frac{Q_cK_c^T}{\sqrt{d}}
   \end{matrix}\right]).
\end{equation}
Here, $\phi(x_{i,j})$ is defined as:
\begin{equation}
    \phi(x_{i,j})=
\left\{
\begin{array}{ll}
   -\infty & \{i,j\}\notin\Omega \\
   x_{i,j} & \text{otherwise}
\end{array}.
\right.
\end{equation}
It is important to note that directly setting the values to $-\infty$ may lead to discontinuous style switching, resulting in an artificial sharp boundary. To achieve a more natural gradient effect, we utilize a linear gradient for $\phi(x_{i,j})$, transitioning from $-\infty$ to $x_{i,j}$. We visually demonstrate the horizontal styling gradient in \Cref{fig:style_grad}, where $\Omega=\{\forall i,j|j>\frac{width}{2}\}$. 

Furthermore, the extension of style transfer from a one-to-one content-style image pair to a one-to-many setting can be easily achieved using Eq.~(\ref{ours_form}). In the case where we aim to transfer the style of $N$ style images to a single content image, the equation is modified as follows:
\begin{equation}
    A'=\sigma(\left[
   \begin{matrix}
       \frac{Q_cK_{s_1}^T}{\sqrt{d}}, & ..., & \frac{Q_cK_{s_N}^T}{\sqrt{d}}, & \frac{Q_cK_c^T}{\sqrt{d}},
   \end{matrix}\right]).
\end{equation}
Owing to the limited memory footprint, we consider this aspect as a potential area for future research and leave it for further investigation.

\begin{figure*}
\setlength{\abovecaptionskip}{2mm}
\centering
\includegraphics[width= \linewidth]{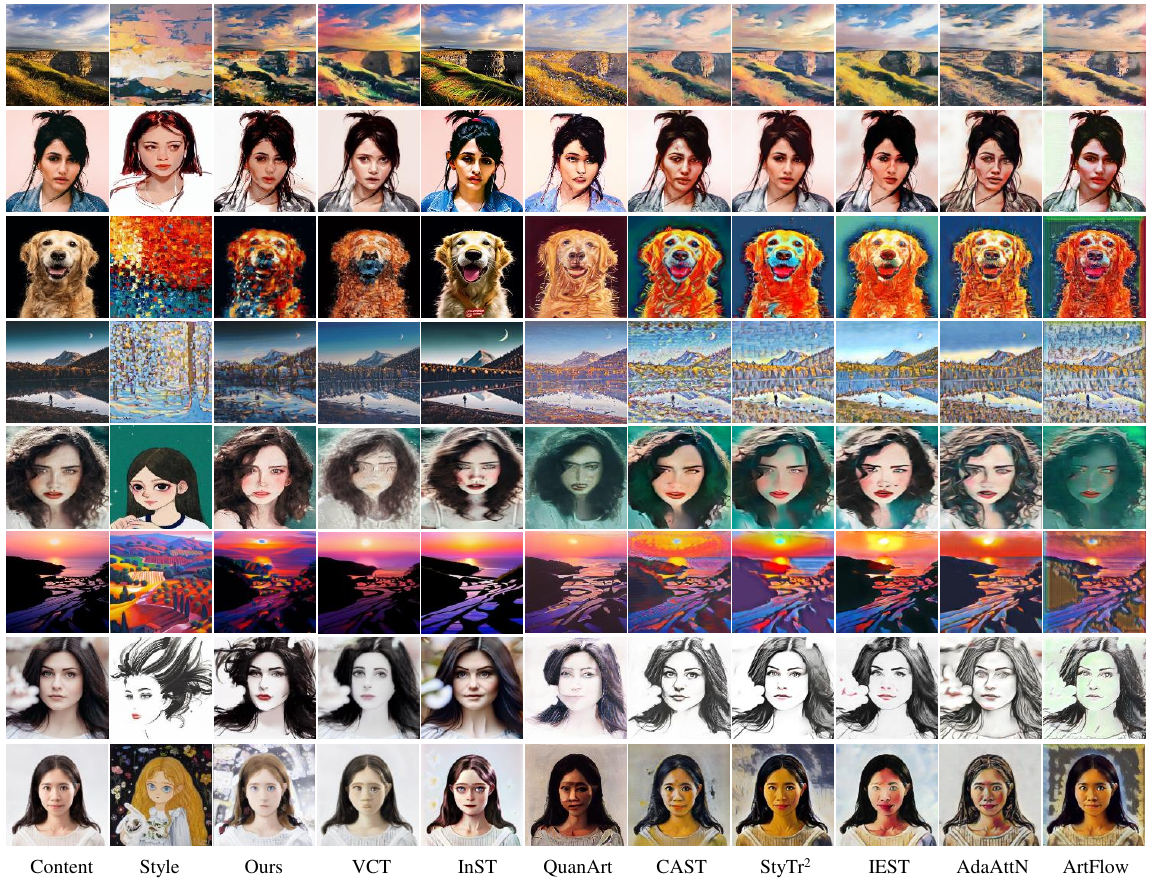}
\caption{Compared with other style transfer methods. The content images are presented in first column, the style images are presented in second column, and the stylized results generated by different methods are presented in the rest. 
}
\vspace{-2mm}
\label{fig:compare}
\end{figure*}

\begin{table*}
\small
\centering
\caption{User study results. Each number represents the percentage of votes that the corresponding method is preferred to ours, using  the criteria of overall quality, preservation of content and style, respectively.}
\resizebox{\linewidth}{!}{
\begin{tabular}{c|ccccccccc}
\toprule
  & VCT~\cite{cheng:2023:ict}& StyleDiff~\cite{jeong:2023:hspace}&InST~\cite{zhang:2023:inversion}&QuanArt~\cite{huang2023quantart}&CAST~\cite{zhang:2022:domain}&StyTr$^2$~\cite{Deng:2022:CVPR} & IEST~\cite{chen:2021:iest} &AdaAttN~\cite{liu:2021:adaattn}&  ArtFlow~\cite{an:2021:artflow}  \\
\midrule
content & 20.6\% & 16.4\%& 18.6\%&35.5\%&23.6\%&45.0\% & 45.9\% & 32.3\%&37.7\% \\
\midrule
style  & 30.7\%  &  25.9\%&  32.3\%&28.6\%&56.4\%&51.4\% & 40.9\% & 40.5\%&23.6\% \\
\midrule
overall & 29.5\% & 16.4\%&  15.9\%&35.9\%&29.5\%&35.5\% & 36.4\% & 35.0\%&22.7\%\\
\bottomrule
\end{tabular}}
\vspace{-10pt}
\label{tab:quancomp}
\end{table*}
\section{Experiments}
\label{sec:Experiments}

\subsection{Implementation Details}
Our research builds upon the concept of Stable Diffusion~\cite{RombachBLE:sd:2O22} and utilizes the v1.5 checkpoint.
In our experimental setup, the text prompts are configured as null character strings. 
We introduce our cross-attention rearrangement module between layers $10$ and $15$. 
The denoising process consists of a total of $30$ steps.

\subsection{Evaluation}

We conduct a comparative analysis of our proposed method against state-of-the-art style transfer approaches, including  ArtFlow~\cite{an:2021:artflow}, AdaAttN~\cite{liu:2021:adaattn}, IEST~\cite{chen:2021:iest}, StyTr$^2$~\cite{Deng:2022:CVPR}, CAST~\cite{zhang:2022:domain}, QuanArt~\cite{huang2023quantart}, InST~\cite{zhang:2023:inversion} and VCT~\cite{cheng:2023:ict}. 
We also compare our method with the domain-specific StyleDiff~\cite{jeong:2023:hspace}, which can only be adapted to pre-trained diffusion models on limited datasets, such as CelebA-HQ~\cite{karras2018progressive}.

\paragraph{Qualitative evaluation.}
The qualitative comparisons presented in Figure~\ref{fig:compare} provide a visual assessment of the outcomes achieved by different style transfer methods. 
ArtFlow~\cite{an:2021:artflow} exhibits limitations regarding style representation and the smoothness of stylization results.
AdaAttN~\cite{liu:2021:adaattn} and IEST~\cite{chen:2021:iest} exhibit inconsistencies in the generated output style compared to the input style reference. 
The results of QuanArt~\cite{huang2023quantart} suffer from a weakening of the style performance due to the prominence of the content appearance. CAST~\cite{zhang:2022:domain} and StyTr$^2$~\cite{Deng:2022:CVPR}  still fall short in terms of faithfully reproducing the artistic qualities, as ``not real compared with artworks'' due to unfaithful style loss constraints. 

Diffusion-based approaches, InST~\cite{zhang:2023:inversion} and VCT~\cite{cheng:2023:ict}, encounter challenges in converging towards the optimal style embedding, leading to failure in generating a similar style and content as the input (as evident from the content deviation in Figure~\ref{fig:teaser} and style deviation in Figure~\ref{fig:compare}). 
VCT~\cite{cheng:2023:ict} demonstrates better style performance, however, still suffers from the issue of content bias, as evidenced by the complete alteration of the girl's facial identity while not precisely matching the desired input style. 
Additionally, it is worth noting that InST~\cite{zhang:2023:inversion} and VCT~\cite{cheng:2023:ict} require a training time of approximately $20$ minutes per style image, whereas our proposed method does not require any training/tuning.
We visualized the results of the comparative analysis with the domain-based method StyleDiff~\cite{jeong:2023:hspace} in the second section of the supplementary materials.
When using images from the CelebA-HQ dataset as content images, the stylized outputs perform well. However, the output becomes confused when going beyond the pre-trained dataset.

In contrast, our proposed approach does not rely on a style loss to enforce conformity of the original image to a different distribution. 
Unlike CNN-based methods, our model does not include a fixed pre-trained encoder, thereby reducing the loss of content and style information. 
As demonstrated in \Cref{fig:compare}, our method achieves captivating stylized results by effectively transferring style patterns, such as painting strokes and lines, onto input content images. These style patterns are skillfully adapted to the content semantics, as exemplified in the \nth{1}, \nth{3}, and \nth{7} rows of \Cref{fig:compare}.


\paragraph{User study.}
To quantitatively evaluate the impact of different stylization methods, we conducted a user study to gather public preferences. 
A total of $55$ participants were randomly selected for the study. 
We provided them with $10$ content images and $10$ style images and generated $100$ stylized results using both our method and comparison methods. 
Each participant was shown $32$ groups of questions, where they were presented with a random content/style image and its corresponding stylized result from both our method and a random comparison method. 
The participants were then asked to answer three questions: 
1) which stylization result better preserves the content, 
2) which stylization result exhibits better style patterns and 
3) which stylization result has a better overall effect.
In total, we collected $1,760$ votes. 
The vote counts are summarized in Table~\ref{tab:quancomp}. 
We can observe that $\mathcal{Z}^*$ beats all contrast methods on average in terms of content preservation.
Only CAST and StyTr$^2$ outperform $\mathcal{Z}^*$ regarding style representation, while the difference is insignificant. 
In terms of the overall effect, $\mathcal{Z}^*$ clearly outperforms the comparative methods; showing that $\mathcal{Z}^*$ achieves a balance between style and content, leading to satisfactory results.
\begin{figure}
\setlength{\abovecaptionskip}{2mm}
\centering
\includegraphics[width= \linewidth]{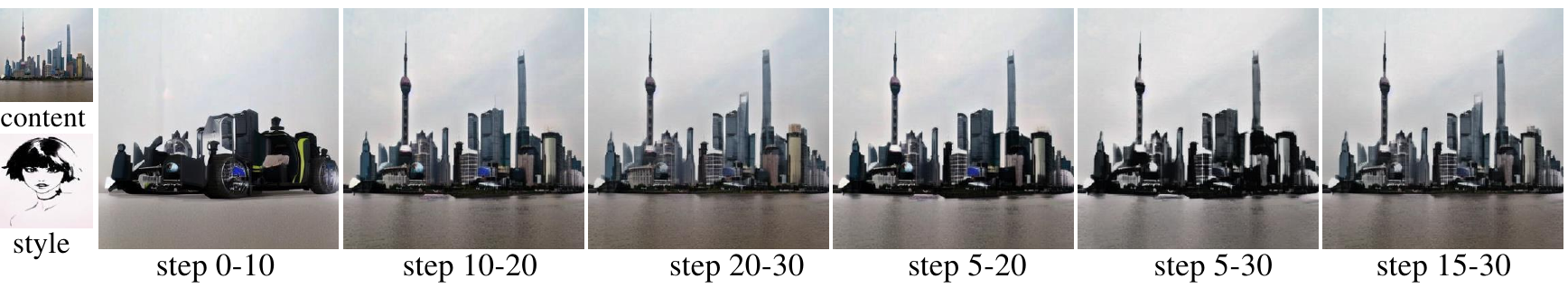}
\caption{Injecting attention module in different denoising steps.
}
\label{fig:a1}
\end{figure}

\begin{figure}
\setlength{\abovecaptionskip}{2mm}
\centering
\includegraphics[width= \linewidth]{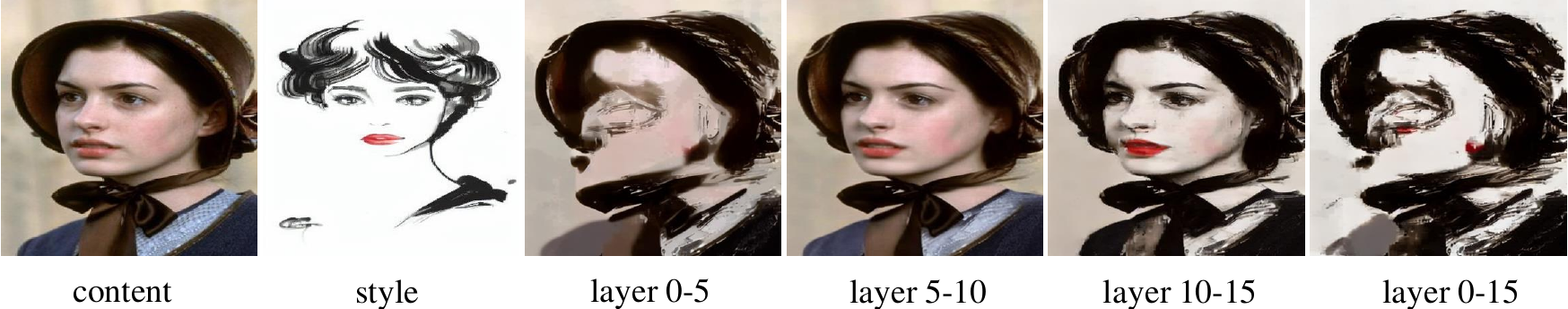}
\caption{Injecting attention module in different U-Net layers.
}
\label{fig:a2}
\end{figure}
\begin{figure}
\setlength{\abovecaptionskip}{2mm}
\centering
\includegraphics[width= \linewidth]{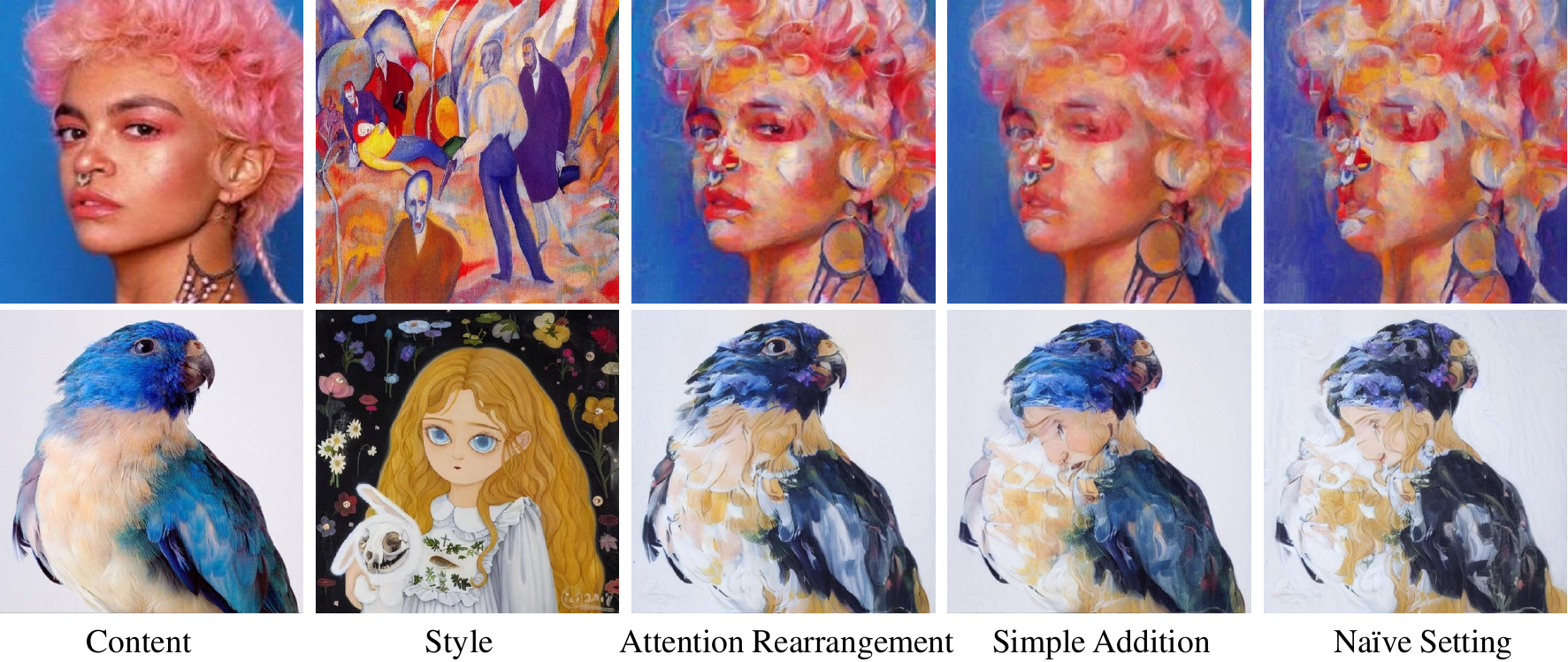}
\caption{Results of using different attention arrangement.
}
\label{fig:attention}
\vspace{-10pt}
\end{figure}



\subsection{Ablation Study}

\paragraph{Influence of the attention injection step and layers.}
The impact of the attention injection step and layers on stylization outcomes is analyzed. 
As depicted in \Cref{fig:a1}, initiating the denoising step too early results in the loss of content structure information, whereas increasing the start denoising step enhances content preservation but sacrifices style patterns (as observed in the first three columns of results). 
On the other hand, increasing the overall denoising steps makes style patterns more prominent without compromising content structure (as evident in the comparison between the second and fourth columns of results).
Consequently, the denoising process is initiated at the \nth{5} step and concluded at the \nth{30} step to achieve the most optimal stylized results.

In \Cref{fig:a2}, the stylized results obtained by injecting the attention module at different layers of the U-Net architecture are presented. 
Utilizing all layers in U-Net adversely affects content structure in the results. 
The low-resolution layers in U-Net effectively preserve content structure but transfer fewer style patterns (layers 5-10). Conversely, the high-resolution layers in U-Net can extract style features (layers 0-5 and layers 10-15), but encoder layers (layers 0-5) damage content structure. 
Hence, the attention module is injected into the high-resolution layers of the decoder.



\paragraph{Influence of the cross-attention rearrangement.}
To evaluate the efficacy of the cross-attention rearrangement, we provide a visual representation of the stylized outcome in \Cref{fig:attention}. 
We conduct a comparative analysis by considering two ablation scenarios: 1) the complete removal of content, as described in Eq.~(\ref{eq:naive_setting}), and 2) the utilization of only the summation operation, denoted by Eq.~(\ref{cross_attention_naive_add}). 
The coefficient $\lambda$ in Eq.~(\ref{cross_attention_naive_add}) is assigned a value of 0.5. 
As illustrated in \Cref{fig:attention}, eliminating the content component leads to compromised preservation of content structures within the stylized outputs. 
Similarly, incorporating solely the Query feature of the content image results in the loss of content details and insufficient style patterns. In contrast, the cross-attention rearrangement demonstrates a superior ability to strike a balance between content and style in the stylized results.

\begin{figure}
\setlength{\abovecaptionskip}{2mm}
\centering
\includegraphics[width= \linewidth]{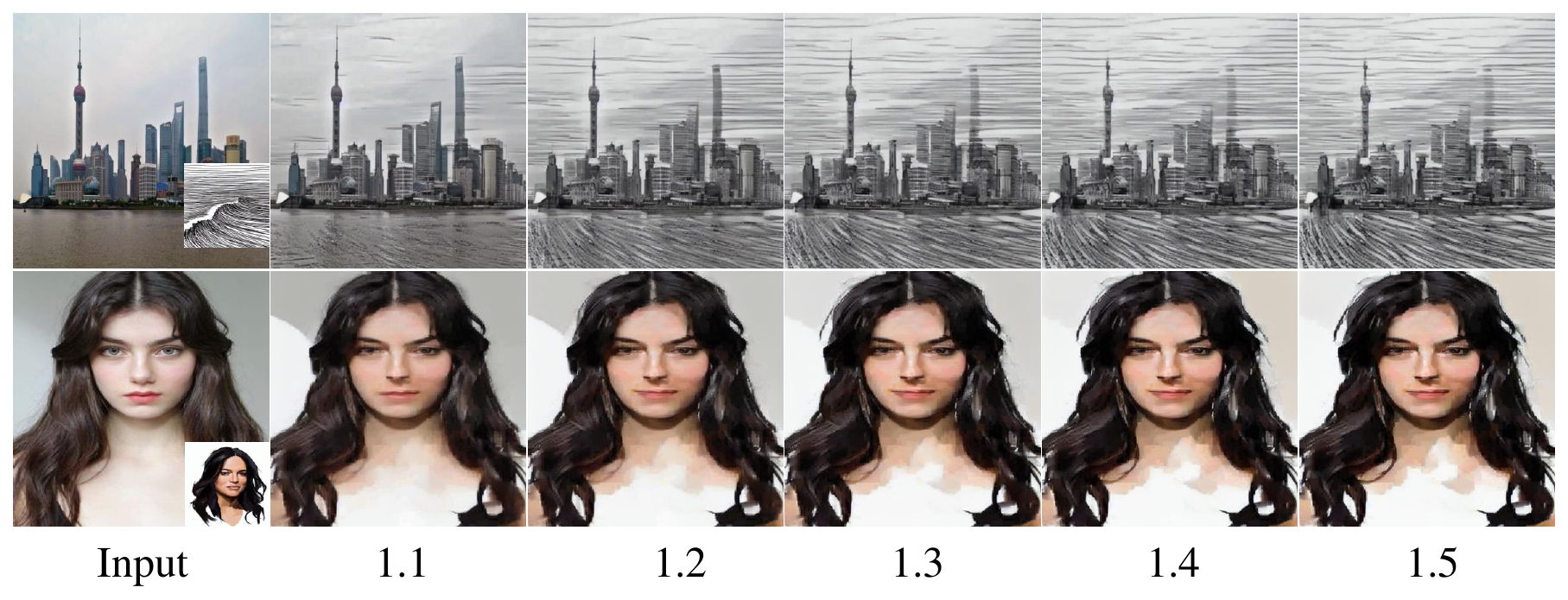}
\caption{The effect of $\lambda$ in Eq.~(\ref{lambda}) .
}
\label{fig:lambda}
\vspace{-10pt}
\end{figure}

\paragraph{Influence of the style scaling factor.}
We present a visualization of the impact of various values of $\lambda$ in Eq.~(\ref{lambda}). 
The results, depicted in Figure~\ref{fig:lambda}, demonstrate the robustness of our method to different choices of $\lambda$. 
Specifically, our approach exhibits satisfactory performance when $\lambda$ is set to values greater than or equal to 1.2. Based on these findings, we adopt a fixed value of $\lambda = 1.2$ for all experiments conducted in this study.

\section{Conclusion}


This paper introduces a novel zero-shot style transfer approach that leverages ample prior knowledge within a pre-trained diffusion model. By incorporating a Key and Value features attention layer, we modify the self-attention mechanism in the diffusion model, enabling the use of Query features to retrieve style-related information from the Key and Value features. To enhance the preservation of content structures in stylized outputs, we propose a cross-attention rearrangement technique that incorporates additional content information and achieves a more favorable balance between content preservation and style rendering. Extensive experimental evaluations demonstrate the superiority of our proposed method in terms of stylization results, outperforming existing state-of-the-art approaches.
{
    \small
    \bibliographystyle{ieeenat_fullname}
    \bibliography{ZST}
}


\end{document}